\documentclass[11pt,a4paper]{article}
\usepackage{times,latexsym}
\usepackage{url}
\usepackage[T1]{fontenc}
\usepackage{times}
\usepackage{latexsym}
\usepackage{multirow}
\usepackage{graphicx}
\usepackage{booktabs}
\usepackage{makecell}

\usepackage{linguex}
\usepackage{hhline}
\usepackage{hyperref}

\usepackage[acceptedWithA]{tacl2021v1}
\usepackage{tacl2021v1}

\usepackage{xspace,mfirstuc,tabulary}

\newcommand{\eat}[1]{}

\newif\iftaclinstructions
\taclinstructionsfalse 
\iftaclinstructions
\renewcommand{\confidential}{}
\renewcommand{\anonsubtext}{(No author info supplied here, for consistency with
TACL-submission anonymization requirements)}
\newcommand{\instr}
\fi

\iftaclpubformat 

\else

\fi

\title{Measure More, Question More: Experimental Studies on Transformer-based Language Models and Complement Coercion}

\author{
Yuling Gu\thanks{ \ \ This is work mainly done when the author was a student at New York University.}\\
Allen Institute for AI, Seattle, WA \\
\texttt{yulingg@allenai.org} 
}

\date{}

\begin{document}
\maketitle
\begin{abstract}
Transformer-based language models have shown strong performance on an array of natural language understanding tasks. However, the question of how these models react to implicit meaning has been largely unexplored. We investigate this using the complement coercion phenomenon, which involves sentences like ``The student finished the book about sailing'' where the action ``reading'' is implicit. We compare LMs' surprisal estimates at various critical sentence regions in sentences with and without implicit meaning. Effects associated with recovering implicit meaning were found at a critical region other than where sentences minimally differ. We then use follow-up experiments to factor out potential confounds, revealing different perspectives that offer a richer and more accurate picture.
\end{abstract}

\section{Introduction}
\label{sec:intro}
While transformer-based language models (LMs) have been evaluated on a range of language understanding tasks, there is much less attention on how they deal with implicit meaning in a phenomenon like complement coercion \citep{pustejovsky1991lexicon, pustejovsky1995lexicon}, where there is an interesting interplay of semantics and commonsense knowledge. In this work, we look into complement coercion in direct object position as a case study for exploring implicit meaning in LMs.

\begin{figure}[t]
  \leftskip -0.3cm
  \includegraphics[width=1.1\linewidth]{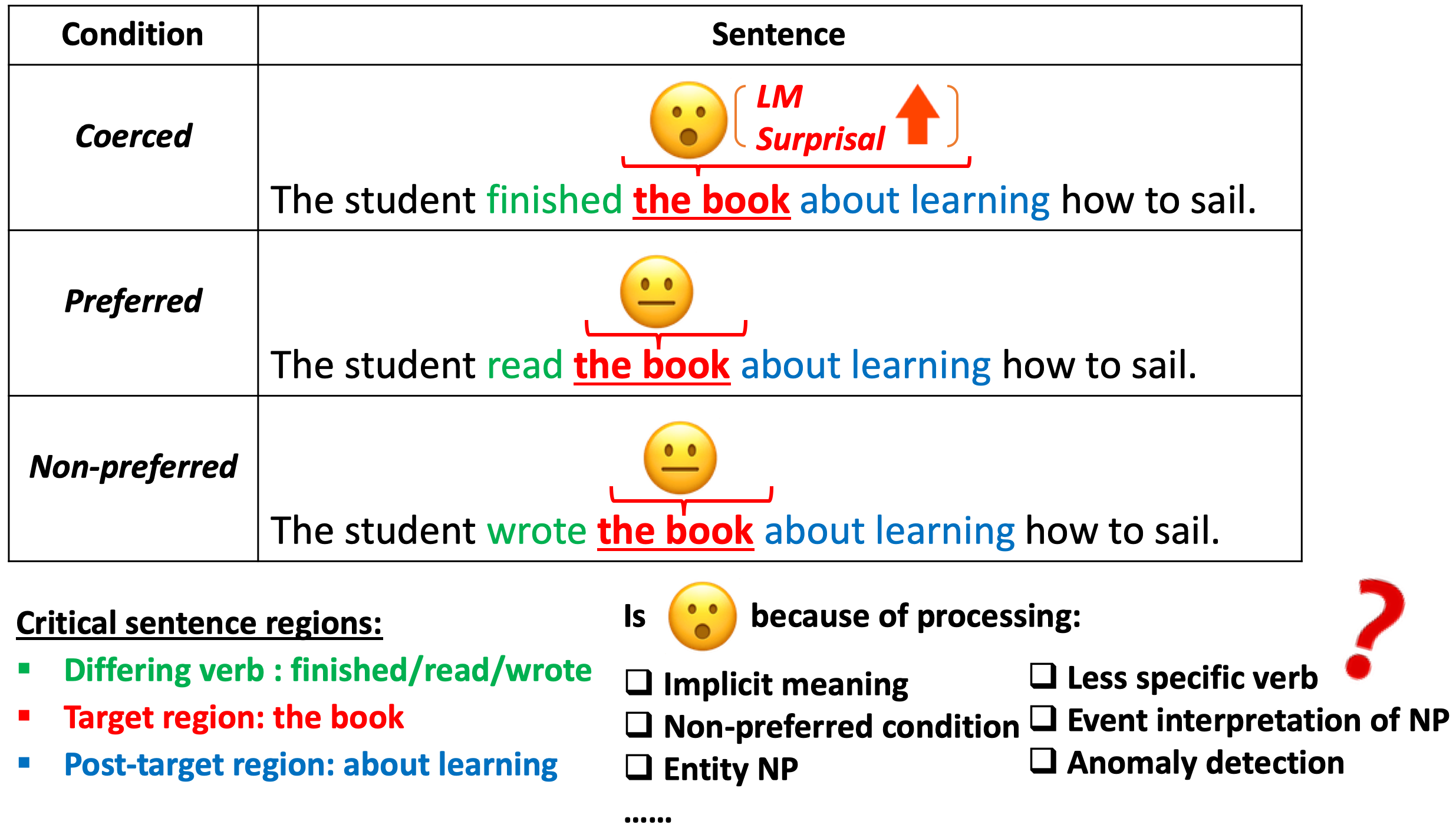}
  \caption{Example of a set of test sentences in experiment 1 and the critical regions for measurement.}
  \label{fig:summary}
\end{figure}

Sentences with complement coercion contain implicit meaning that we infer based on the semantic meaning of the actions or objects in the sentence as well as our commonsense knowledge about them. Consider the sentence in the coerced condition in Figure \ref{fig:summary}, one would expect an event-selecting verb like \textit{finished} to be followed by a complement that expresses some event or activity rather than a complement phrase like \textit{the book} which has the default interpretation of an entity. Interestingly, although there is no mention of the initiated activity, the meaning that the student began \textit{some activity} involving the book is still available and a reader would interpret the sentence such as in the second (preferred) condition in Figure \ref{fig:summary}, a sentence which involves no implicit meaning.

In NLP, a fundamental question as to \textit{how} LMs behave as they progress through such sentences with implicit meaning in their processing remains unanswered, and it is not yet clear whether they behave differently for sentences with and without implicit meaning. This paper aims to improve our understanding of how neural LMs behave when processing English sentences with and without implicit meaning by using the psycholinguistic phenomenon of complement coercion (Section \ref{sec:coercion}). Using the GPT-2 \citep{radford2019language} family models as a case study, we  use surprisal estimates at critical sentence regions to model LMs' behavior (see Figure \ref{fig:summary}, details in Section \ref{sec:experiment}). We focus on complement coercion in the direct object position, introducing a series of three diagnostic datasets drawn from human psycholinguistics experiments (Section \ref{sec:dataset}). Given their origin in psycholinguistics, these diagnostic datasets are controlled to ask targeted questions regarding the processing of implicit meaning and specifically chosen to elicit differences in human processing of coercion versus control conditions.

This paper has three main contributions. First, to the best of our knowledge, our work is the first of its kind to study transformer-based LMs' behavior on the psycholinguistic phenomenon of complement coercion in English, using surprisal estimates. Second, while most previous works studying LMs' behavior compare full sentences \cite{warstadt-etal-2020-blimp-benchmark} or examine one critical region per sentence \cite{hu-etal-2020-systematic}, for each sentence, we take measurements at three positions important for analysis of the complement coercion phenomenon to provide a richer analysis. Finally, the series of three carefully designed experiments we present in the paper provides an illustrative example of how targeted follow-up experiments could be effectively used to tease apart confounding factors in probing experiments to give us a more accurate picture. Our first experiment provides strong evidence that GPT-2 models consistently distinguish sentences with implicit meaning (coerced conditions) from those without. However, rigorous follow-up experiments reveal that this apparent distinction between the two conditions can be completely attributed to confounding factors.

\section{What is coercion?}
\label{sec:coercion}

In natural language, the meaning of an expression is often thought to be constructed from the meaning of its smaller parts \cite{fodor2002compositionality, heim1998generativeGrammar, montague1970univGrammar}. However, there are also many cases where the meaning of an expression may be richer than what is overtly expressed and composed from its constituents, giving rise to various semantic theories on non-syntactic operations that contribute extrasyntactic content to meaning \cite{groenendijk1989interrogatives, hendriks1998quantificationCoord, partee1986NPshift, partee1995semanticsComp, partee1983conjTypeAmb}. One such process of semantic enrichment is through coercion, where a linguistic element that has a default interpretation mismatched with the semantic context is coerced to take on another meaning that is more compatible semantically \cite{jackendoff1997archLang, pustejovsky1995lexicon}. We focus our study on complement coercion in the direct object position which we discuss in detail in Section \ref{subsec:complDO}.

\subsection{Complement coercion in direct object position}
\label{subsec:complDO}
Complement coercion occurs when a verb, such as \textit{begin, finish, complete}, that semantically selects (s-selects) for event-describing complements, takes on a complement that has the default interpretation of an entity instead. One syntactic environment that complement coercion can occur in is the direct object position, when a verb like \textit{finished} is composed directly with an entity-denoting determiner phrase (NP) in direct object position \citep{traxler2002Reading, mcElree2006enrichedComp, pylkkanen2007silentMeaning, pylkkanen2008mismatch} as in the coerced condition in Figure \ref{fig:summary}.

In the coerced condition in Figure \ref{fig:summary}, \textit{finished} s-selects for an event-describing complement and is most naturally combined with a VP-complement. However, the NP \textit{the book} has the default interpretation of an entity. As such, the composition of the verb \textit{finished} with the entity-denoting NP \textit{the book} presents a ``type-mismatch'' \citep{partee1986NPshift}. But instead of resulting in an anomalous sentence, this sentence very naturally received the interpretation that the student finished some contextually relevant, though not overtly stated, activity involving the book. Therefore, the semantic composition of the coerced condition in Figure \ref{fig:summary} involves some element that is not present in its syntax. Due to the enriched composition, readers typically interpret that sentence the same way as one that explicitly expresses the meaning of the event (like the preferred condition in Figure \ref{fig:summary}).

In the reconstruction of the covert event, semantic knowledge that \textit{finished} s-selects for an event-describing complement sets up the expectation that it will be followed up actions like \textit{reading, writing, eating, watching, etc}.  We then rely on our commonsense knowledge, that someone cannot be ``eating'' or ``watching'' \textit{the book} to rule these possibilities out. Finally, the subject, \textit{the student}, together with our world-knowledge that it is more common for a \textit{student} to read books rather than to write them, helps us to infer ``reading'' as the implied action. Therefore, interpreting the implicit meaning here involves both semantics and commonsense knowledge.

\subsection{Coercion in cognitive sciences}

The property of natural language that the meaning of certain expressions cannot be recovered from what has been explicitly expressed has garnered much attention both in the linguistics and psycholinguistics community. Some variants of coercion have been studied from a linguistics perspective looking into the theories behind this phenomenon using concepts such as type-shifting, contributing to a rich literature including \citet{barker2016CQnominals, johannes1992sort, groenendijk1989interrogatives, heim1998generativeGrammar, jackendoff1997archLang, partee1983conjTypeAmb}. The underlying shifting operation which coerces the entity-denoting NP from its default interpretation into an event to satisfy the selectional properties of the verb (discussed in Section \ref{subsec:complDO}), has also received much attention in linguistics \cite{partee1986NPshift, partee1995semanticsComp, jackendoff1997archLang, pustejovsky1991lexicon, pustejovsky1993typelexicon, pustejovsky1995lexicon}. 

In the psycholinguistics community, there has been an increasing body of work reaffirming the significance of the phenomenon using experiments measuring eye-movements \citep{traxler2002Reading, harris2008eyeTracking, mcElree2006dickens}, reading times \citep{traxler2002Reading, mcElree2001ReadingTE, pylkkanen2008mismatch, pylkkanen2009aspCoer, mcElree2006enrichedComp} and magnetoencephalography (MEG) studies \citep{harris2008eyeTracking, pylkkanen2007silentMeaning, pylkkanen2008mismatch, pylkkanen2009aspCoer,pylkkanen2009adjCoer}. Specifically, for the case of complement coercion in direct object position,
 \citet{mcElree2001ReadingTE, traxler2002Reading} and  \citet{mcElree2006enrichedComp} found increased cost of processing associated with coerced conditions in terms of reading time for complements; \citet{traxler2002Reading, traxler2005Reading} and \citet{mcElree2006dickens}'s results through tracking eye-movements also point in the same direction; \cite{pylkkanen2007silentMeaning} localized the cost of such processing by observing increased amplitudes in the anterior midline field (AMF) in ventromedial prefrontal areas at 350-450 msec.

However, the question of how transformer-based LMs handle implicit meaning in the phenomenon of coercion has received much less attention. To the best of our knowledge, our work is the first to provide a systematic study on this.

\section{Experiment}
\label{sec:experiment}
To study transformer-based LMs' behavior, we use the conditional probability (subsection \ref{subsec:experiment_LM}) of auto-regressive LMs to compute surprisal estimates (subsection \ref{subsec:experiment_surprisal}), and compare surprisal estimates at critical sentence regions (subsection \ref{subsec:experiment_measure_pos}). Given that GPT-2 \citep{radford2019language} models have been shown to better predict human neural and behavioral data during language processing compared to other LMs \citep{schrimpf-2021-predictive, Goldstein2022SharedCP}, we focus on the family of GPT-2 models as a case study (subsection \ref{subsec:experiment_models}). 

\subsection{Language models' output probabilities}
\label{subsec:experiment_LM}

Let $S$ be a sentence with tokens $w_1 ,\cdots, w_N$, the conditional probability at token $i$, $P(w_i|w_1,\cdots, w_{i-1})$ is estimated by the model either using statistical-based methods like $n$-gram \citep{chen1999empirical} or neural network-based methods like LSTM \citep{mikolov2010recurrent} and transformer \citep{radford2019language}. The conditional probability from auto-regressive LMs only depends on the words to the left of that position. That is, when an auto-regressive LM predicts the token at position $i$, it only relies on tokens $w_1 ,\cdots, w_{i-1}$, and we use this to model human's processing of token $i$ in a sentence having read what came before it. To obtain the conditional probability, we apply softmax to the logits produced by the LMs. This probability is then used to compute the surprisal score.

\subsection{Surprisal}
\label{subsec:experiment_surprisal}
\citet{hale-2001-probabilistic} and \citet{LEVY2008expectationbased}
suggested a quantification of the cognitive effort required to process a word in a sentence using the surprisal of the word: 
\[
S(w_i)= -\log_{2} P(w_i|w_1,\cdots, w_{i-1}). 
\]
In previous work, \citet{Schijndel2018ModelingGP}
used surprisal estimates from LMs (RNN models with LSTM units) to model increased processing difficulty associated with increased reading times in garden path sentences and \citet{Demberg2008DataFE}
showed that surprisal from a broad-coverage unlexicalized (i.e., only using the structural probabilities) LM strongly correlates with reading times. Looking at the phenomenon of coercion in German, \citet{DELOGU2017CoercionSurprisalERP} presents results from monitoring eye-tracking and event-related potentials (ERPs) as well as estimating the surprisal of the complement noun from a trigram language model, showing that the coercion cost could be largely attributed to the surprisal of the complement noun. 

In this work, we look at the phenomenon of coercion in English, and compute the surprisal estimates for transformer-based LMs to analyze the LMs' behavior in processing sentences. Surprisal estimate for each region is calculated by averaging the surprisal estimates of all words in that region. We follow \citet{LEVY2008expectationbased}'s original proposal of the surprisal theory in interpreting surprisal as going beyond the traditional domain of predictability (primarily semantic phenomenon) and that it can arise from any source.

\subsection{Measure positions}
\label{subsec:experiment_measure_pos}
\begin{table*}[]
    \small
    \centering
    \begin{tabular}{c|c|c}
    \hline
     \makecell{Measured\\position} & What it comprises &  \makecell{Illustration of the position in example set} \\
     \hline
     \multirow{3}{*}{\makecell{Differing\\verb}} & \multirow{3}{*}{\makecell{Verb which sentences in \\each set minimally differ}} & Coerced: The secretary \textbf{began} the memo about the new office policy. \\
     & & Preferred: The secretary \textbf{typed} the memo about the new office policy. \\
     & & Non-preferred: The secretary \textbf{read} the memo about the new office policy. \\
     \hline
     \multirow{3}{*}{\makecell{Target\\region}} & \multirow{3}{*}{\makecell{ Determiner and noun \\ following the differing verb}} & Coerced: The secretary began \textbf{the memo} about the new office policy. \\
     & & Preferred: The secretary typed \textbf{the memo} about the new office policy. \\
     & & Non-preferred: The secretary read \textbf{the memo} about the new office policy. \\
     \hline
    \multirow{3}{*}{\makecell{Post-target\\region}} & \multirow{3}{*}{\makecell{Two words immediately \\ following the target region}} & Coerced: The secretary began the memo \textbf{about the} new office policy. \\
     & & Preferred: The secretary typed the memo \textbf{about the} new office policy. \\
     & & Non-preferred: The secretary read the memo \textbf{about the} new office policy. \\
    \hline
 
    \end{tabular}
    \caption{Details of critical sentence regions measured for analyzing LMs. Illustrated examples are based on a triplet in our diagnostic Dataset 1 as found in \citet{traxler2002Reading}.}
    \label{tab:three_measure_positions}
\end{table*}

In a phenomenon like coercion, the effects associated with the coerced condition are not uniform across every word in a sentence and can even vary across regions where the effects may show up. Therefore, it is inadequate to provide an analysis based on making one measurement per sentence or simply averaging across each sentence. Based on insights from psycholinguistic experiments like \citet{traxler2002Reading}, we capitalize on making measurements at three critical sentence regions important for analysis of the coercion phenomenon to provide a richer analysis. These regions are summarized in Table \ref{tab:three_measure_positions} with illustrative examples. First, we pay attention to the verb region of each sentence in which the sentence differs from one or more of the other sentences in the set (triplet or quadruplet). Next, we examine the noun and its determiner following the verb region as the target region. Coerced conditions incur additional processing cost when readers realize that the default interpretation of the entity-denoting NP does not match the selectional restrictions of the preceding verb, and this happens when (or shortly after) readers encounter the NP \citep{traxler2002Reading}, making the NP an important region to analyze. Finally, as effects sometimes emerge on words following the target region \citep{Rayner1989}, we also examine the post-target region which comprises the two words following the target region.

\subsection{Models}
\label{subsec:experiment_models}
\begin{table}
    \small\addtolength{\tabcolsep}{-3pt}
    \begin{tabular}{c|c|c|c|c}
    \hline
    Model & Layers & Hidden & Heads & Parameters \\
    \hline
    DistilGPT-2 & 6 & 768 & 12 & 82M \\
    GPT-2 & 12 & 768 & 12 & 117M \\
    GPT-2 medium & 24 & 1024 & 16 & 345M \\
    GPT-2 large & 36 & 1280 & 20 & 774M \\
    GPT-2 xl & 48 &  1600 & 25 & 1558M \\
    \hline
    \end{tabular}
    \caption{Details of the five GPT-2 models studied.}
    \label{tab:model_details}
\end{table}
GPT-2 uses multi-layer transformer decoder as the architecture, and its pretraining uses left-to-right autoregressive language modeling. We compare results across four LMs of varying sizes as introduced in the original GPT-2 paper \citep{radford2019language}, together with DistilGPT-2, distilled from the base GPT2 model using the same distillation method as in \citet{Sanh2019DistilBERTAD}. The architectures of the models used are summarized in Table \ref{tab:model_details}. For all models analyzed in this paper, we use the Huggingface implementation \citep{Wolf2019HuggingFacesTS, wolf-etal-2020-transformers},\footnote{\url{https://github.com/huggingface/transformers}} based on PyTorch \citep{Paszke2019pytorch}. Following \citet{Ettinger2019WhatBI} and \citet{ hu-etal-2020-systematic}, since our diagnostic tests are designed to study processing of words in context (rather than performance on a particular task),  we examine the models' behavior without task-specific fine-tuning.

\section{Diagnostic datasets}
\label{sec:dataset}
\begin{table*}[h]
    \centering
    \begin{tabular}{c|c|c}
    \toprule
     Dataset & Original psycholinguistic experiment  & Selection from original stimuli\\
     \hline
     1  & \multirow{2}{*}{\makecell{``Coercion in sentence processing: \\Evidence from eye-movements and \\self-paced reading'' \\ by \citet{traxler2002Reading}}} & \makecell{36 triplets \\ ( Coercion/Preferred/Non-preferred) \\ from stimuli for Experiment 1}\\
     \hhline{-~-}
    2  &  & \makecell{32 quadruplets (Event/Neutral verb + \\ Event/Entity NP) from 
    \\stimuli for Experiments 2 and 3 \\ }\\
    \hline
    3 & \makecell{``An MEG Study of Silent Meaning'' \\ by \citet{pylkkanen2007silentMeaning}} & \makecell{35 triplets \\ (Coerced/Anomalous/Control) \\ from the \textit{Nonembedded Stimuli}} \\
    \bottomrule
    \end{tabular}
    \caption{Source of diagnostic datasets used in our experiments.}
    \label{tab:coercion_source}
\end{table*}

For our diagnostic datasets, we use expert-designed sentence sets from psycholinguistic experiments as summarized in Table \ref{tab:coercion_source}. Each triplet or quadruplet contains one sentence with the coerced condition, and other sentences specifically designed to tease apart different potential confounding factors which we will elaborate on in the rest of this section.

\subsection{Dataset 1: Coercion/Preferred/Non-preferred}
\label{dataset: dataset1}
\subsubsection{Targeted design}
Our first dataset is targeted at investigating if LMs show differences in their behavior for coerced versus non-coerced conditions. Further, it is designed to answer the potential follow-up question of whether we can rule out that the observed differences in their behavior are due to differing levels of preference for the conditions rather than associated with recovering the implicit meaning in coerced condition. This dataset originates from Experiment 1 performed in \citet{traxler2002Reading} which compared measurements of eye-movements, showing that humans demonstrate rapid difficulty soon after encountering the NP complement when verbs that require an event are followed by complements typically used as entities rather than events. The data comprises 36 triplets, and we illustrate the format of each of those triples, by showing an example in \ref{traxler_exp1} a-c.

\ex. \label{traxler_exp1}
\a. \label{traxler_exp1_a} \textbf{Coerced}: The secretary began the memo about the new office policy.
\b. \label{traxler_exp1_b} \textbf{Preferred}: The secretary typed the memo about the new office policy.
\c. \label{traxler_exp1_c} \textbf{Non-preferred}: The secretary read the memo about the new office policy.

Sentences like \ref{traxler_exp1_a} present coerced conditions which involve implicit meaning as readers need to reconstruct the covert event. In contrast, in sentences \ref{traxler_exp1_b} and \ref{traxler_exp1_c}, the verb explicitly specifies the activity. The difference between sentences \ref{traxler_exp1_b} and \ref{traxler_exp1_c} is that according to a preference norming pre-test, sentences like \ref{traxler_exp1_b} present the preferred interpretation of sentences such as \ref{traxler_exp1_a}, whereas sentences like \ref{traxler_exp1_c} present a plausible but non-preferred interpretation. To derive the preferred and non-preferred conditions, \citet{traxler2002Reading} collected data on how participants would typically interpret sentences through fill-in-the-blank tasks involving type-shifting sentences like ``The editor finished \rule{1cm}{0.15mm} the article.''

\subsubsection{Studying LMs with the dataset}
If LMs show differences in their behavior for coerced versus non-coerced conditions representing the extra processing cost, surprisal measured at some measure position would be greater for condition (a) in example \ref{traxler_exp1} compared to conditions (b) and (c). To further differentiate if this difference is due to the additional processing associated with reconstructing the covert event rather than processing atypical non-preferred relationships, it is important that this difference is not significant when comparing the preferred condition (b) to the non-preferred condition (c). This would support that the coerced condition incurs additional processing cost in reconstructing the implicit meaning and rule out the effect of costs associated with processing non-preferred conditions.

\subsection{Dataset 2: Event/Neutral verb + Event/Entity NP}
\label{dataset: dataset2}
\subsubsection{Targeted design}
Our second dataset is targeted to investigate if we can rule out two potential alternative explanations involving elements in the coerced conditions -- whether the potential differences in LMs' behavior for coerced versus non-coerced conditions are in fact due to (1) event-selecting verbs like \textit{began} being simply semantically underspecified, or (2) event interpretations of NPs being more difficult to compute, causing different behavior associated with general processing difficulty rather than one associated with reconstructing the implicit meaning. This dataset originates from Experiments 2 and 3 performed in \citet{traxler2002Reading} which compared measurements of eye-movements and self-paced reading respectively. \citet{traxler2002Reading}'s experimental results showed that neither of the two factors affected processing difficulty, ruling out the two alternative explanations for the differences in processing observed in humans. This data comprises 32 quadruplets, and we illustrate the format of each of those quadruplets, by showing an example in \ref{traxler_exp23} a-d.

\ex. \label{traxler_exp23}
\a. \label{traxler_exp23_a} \textbf{Event verb + Event NP}: The boy started the fight after school today.
\b. \label{traxler_exp23_b} \textbf{Neutral verb + Event NP}: The boy saw the fight after school today.
\c. \label{traxler_exp23_c} \textbf{Event verb + Entity NP}: The boy started the puzzle after school today.
\d. \label{traxler_exp23_d} \textbf{Neutral verb + Entity NP}: The boy saw the puzzle after school today.

Sentences like \ref{traxler_exp23_c} present coerced conditions because with event-selecting verbs like \textit{started}, the entity-denoting NP complement \textit{the puzzle} requires type-shifting to give an event interpretation and readers need to reconstruct the implicit action (e.g. \textit{solving}). Sentences of the form in \ref{traxler_exp23_b} and \ref{traxler_exp23_d}, with neutral verbs like \textit{saw} which can be used felicitously with both entity-denoting or event-denoting complements, do not require such coercion operation. Sentence \ref{traxler_exp23_a} also presents a non-coerced condition since the event-denoting complement \textit{the fight} satisfies the selectional restrictions of the verb.

\subsubsection{Studying LMs with the dataset}
If LMs show differences in their behavior for coerced versus non-coerced conditions, in at least one of the measure positions, they would have the greatest surprisal in for coerced condition (c) in example \ref{traxler_exp23} compared to the other three conditions. However, if this greater surprisal is only due to verbs like \textit{started} being (arguably) less specific than verbs like \textit{saw}, behavior within condition pairs (a) and (c), as well as (b) and (d) would be similar, and any difference would only be across the pairs where the surprisal for (a) and (c) would be greater than that of (b) and (d). For the second confounding factor, if the difference is generally associated with processing event interpretations of NPs rather than computing implicit meaning, LMs' surprisal on conditions (a), (b) and (c) where NPs are assigned event interpretations would each be significantly greater than condition (d) where the NP takes on an entity interpretation. We use this diagnostic dataset to tease apart these two possible confounding factors.

\subsection{Dataset 3: Coerced/Anomalous/Control}
\label{dataset: dataset3}
\subsubsection{Targeted design}

Our final diagnostic dataset is designed to answer the follow-up question of whether we can rule out that any observed effect of coercion the LMs show is due to treating the coerced condition as an anomalous condition (when it is in fact fully grammatical and plausible) rather than associated with recovering the implicit meaning. This dataset originates from the \textit{Nonembedded Stimuli} used in \citet{pylkkanen2007silentMeaning}, which compared MEG measurements, showing that the coerced and anomalous conditions elicited distinct effects, ruling out the possibility that the coerced condition is processed like an anomaly in reading. The data comprises 35 triplets, and we illustrate the format of those triplets by showing an example in \ref{pylkkanen_silentMeaning_exp} a-c.

\ex. \label{pylkkanen_silentMeaning_exp}
\a. \label{pylkkanen_silentMeaning_a} \textbf{Coerced}: The journalist began the article before his coffee break.
\b. \label{pylkkanen_silentMeaning_exp_b} \textbf{Anomalous}: The journalist astonished the article before his coffee break.
\c. \label{pylkkanen_silentMeaning_exp_c} \textbf{Control}: The journalist wrote the article before his coffee break.

Sentences like \ref{pylkkanen_silentMeaning_a} present coerced conditions which require the reader to posit an activity, relate the entity-denoting NP to that activity, and construe the activity as the complement of the event-selecting verb. In contrast, similar to the verbs in preferred and non-preferred conditions for Dataset 1, as well the neutral verbs in Dataset 2, the verb in \ref{pylkkanen_silentMeaning_exp_c} explicitly expresses an eventive interpretation in which readers do not need to generate the eventive meaning that is not part of the overt linguistic input. An anomalous condition like \ref{pylkkanen_silentMeaning_exp_b} is also introduced in each set of sentences. The anomalous sentence is constructed with verbs like \textit{astonish} which require direct objects that denote an experiencer of the psychological state. This requirement cannot be fulfilled with entity-denoting NPs such as \textit{the article}, resulting in an anomaly. 

\subsubsection{Studying LMs with the dataset}
Similar to the case for Dataset 1, if LMs show differences in their behavior for coerced versus non-coerced conditions representing the extra processing cost, there will be greater surprisal at some measure position for condition (a) compared to (c). Further, if this effect is different from anomaly detection, the observed effects for condition (a) will be different from that of condition (b).

\section{Results and analysis}
\begin{figure*}
  \includegraphics[width=\textwidth]{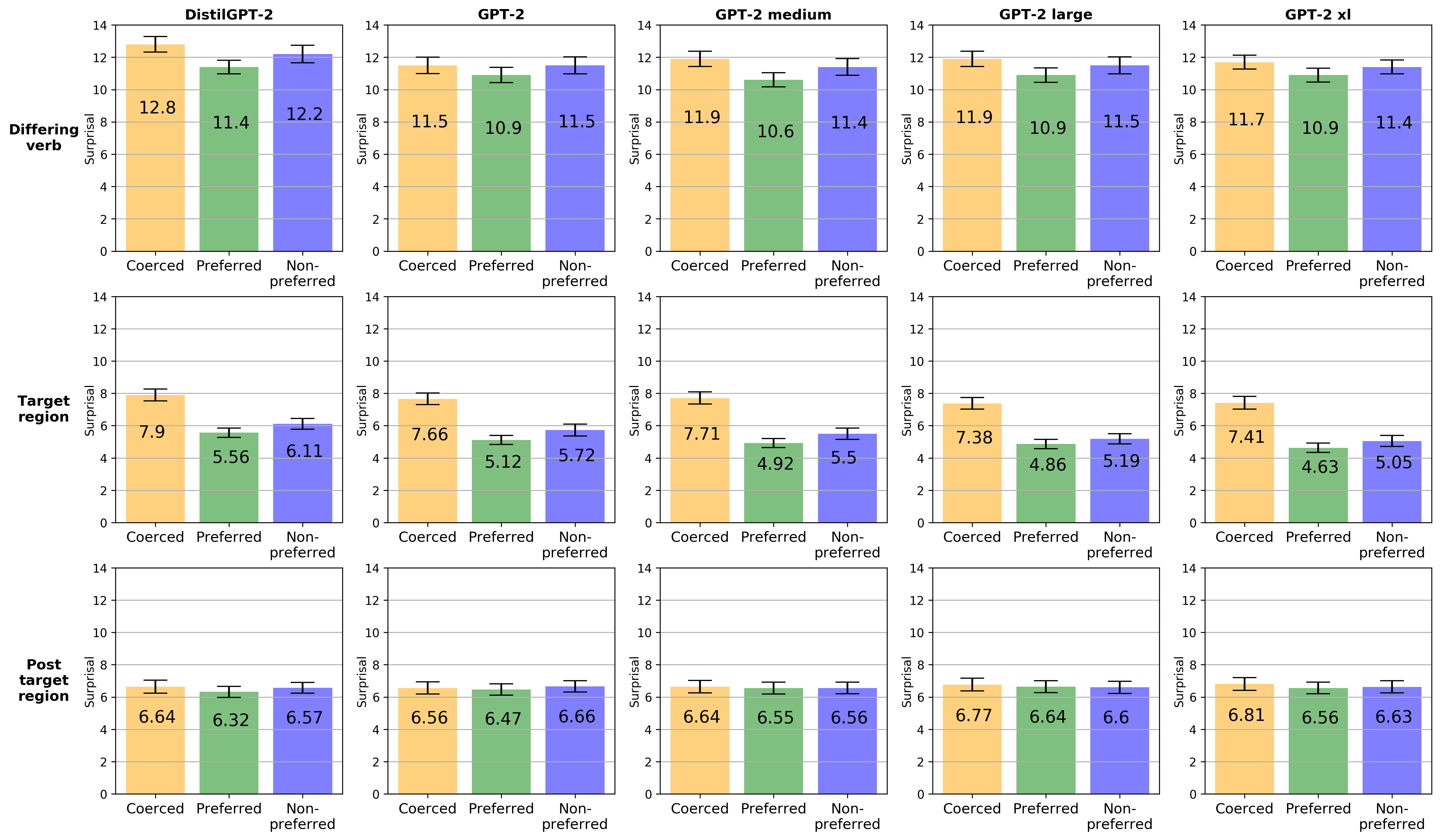}
  \caption{Bar graphs showing mean surprisal estimates from Experiment 1,  by model (column), region (row) and condition (x-axis of each subplot). The error bars represent standard error. At the target region (see corresponding row), there is consistently greater surprisal in the coerced condition (first bar in each subplot) than other conditions.}
  \label{fig:exp1}
\end{figure*}

\begin{figure*}
  \includegraphics[width=\textwidth]{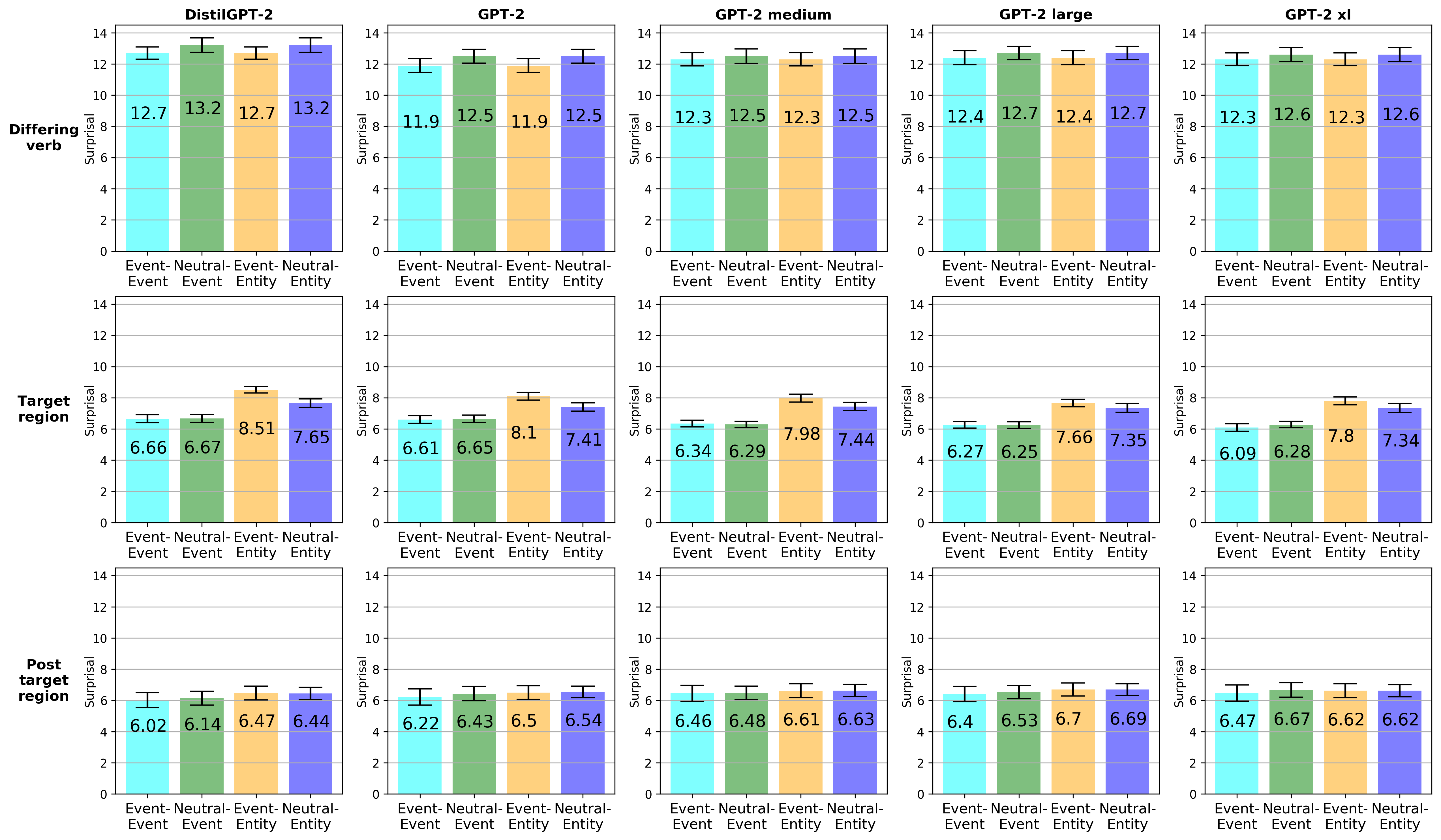}
  \caption{Bar graphs showing mean surprisal estimates from Experiment 2,  by model (column), region (row) and condition (x-axis of each subplot). The error bars represent standard error. At the target region (see corresponding row), there is generally significantly greater surprisal for sentences with entity-denoting NPs (two rightmost bars in each subplot).}
  \label{fig:exp2}
\end{figure*}

\begin{figure*}
  \includegraphics[width=\textwidth]{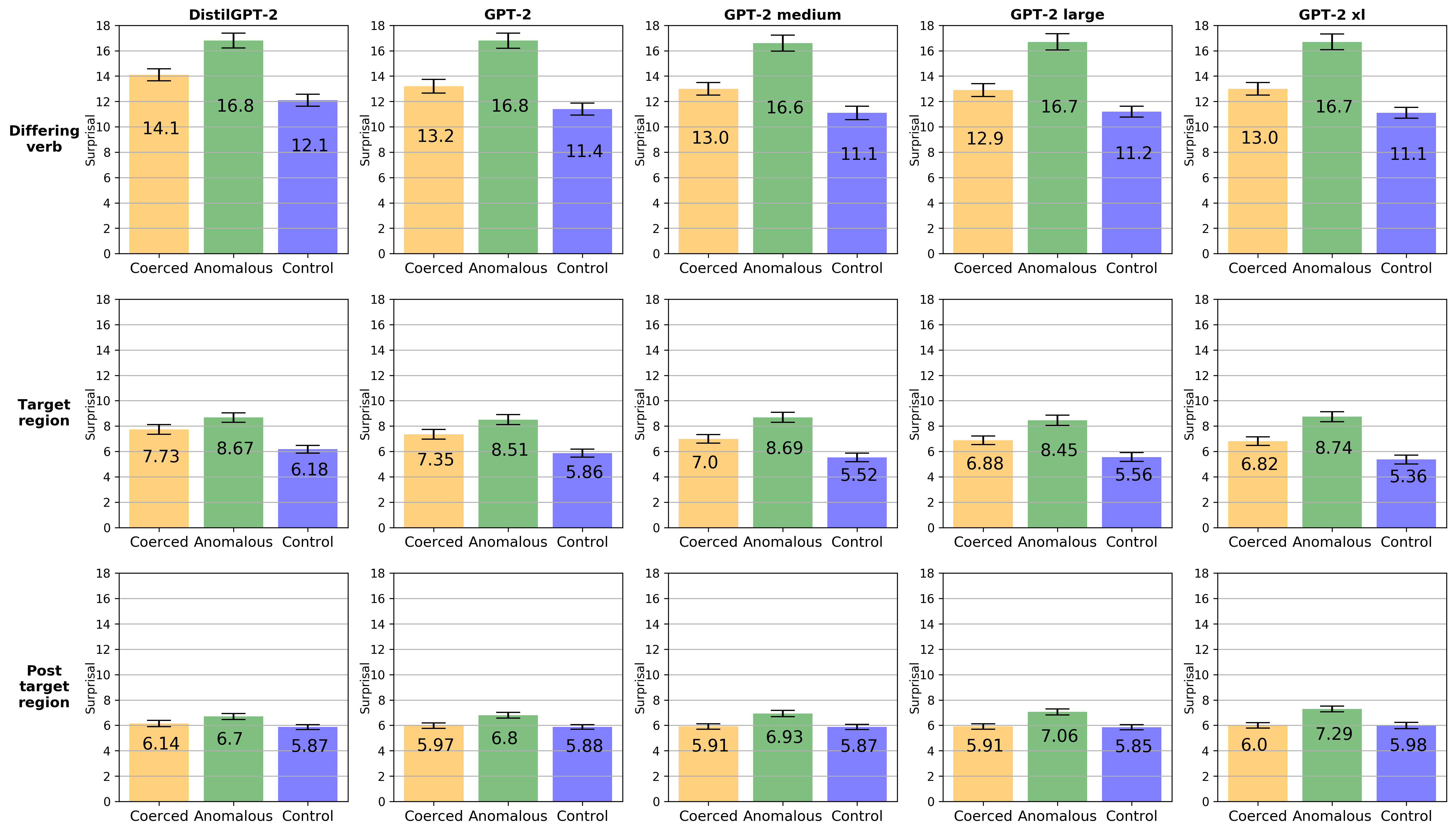}
  \caption{Bar graphs showing mean surprisal estimates from Experiment 3,  by model (column), region (row) and condition (x-axis of each subplot). The error bars represent standard error. Across all five models and all three measure positions,  there is consistently greater surprisal for processing the anomalous condition (second bar in each subplot) than the coerced condition (and control condition).}
  \label{fig:exp3}
\end{figure*}

For the different diagnostic datasets and LMs experimented, we report surprisal estimates across the three critical sentence regions. The surprisal estimates from measurements on the three datasets are presented in Figures \ref{fig:exp1}, \ref{fig:exp2} and \ref{fig:exp3} respectively. To test for statistical significance, we performed Friedman test for repeated measures and post-hoc Wilcoxon matched-pairs signed-rank test with Bonferroni correction. We report differences that are significant under both the Friedman test and post-hoc test. An alpha level of $0.05$ was used.

\subsection{Experiment 1: Coerced/Preferred/Non-preferred}
It is consistent across all five models that at the target region, Friedman's test of differences among repeated measures showed that there is a significant difference (in all cases, $p < 0.001$) among surprisal estimates measured in the coerced, preferred and non-preferred conditions. 
\label{result: dataset1}
\subsubsection{Greater surprisal in coerced condition}
For all five models, post-hoc tests using Wilcoxon matched-pairs signed-rank test with Bonferroni correction showed that surprisal for the target region in the coerced condition is significantly greater than both that in the preferred and non-preferred conditions ($p < 0.001$ for all pairwise comparisons). This result suggests that surprisal estimates in LMs reflect the difficulty associated with recovering the covert meaning in the coerced condition compared to other conditions.

\subsubsection{Greater surprisal $\neq$ non-preferred}
Further, based on post-hoc tests for surprisal estimates at the target region, it is also consistent across the LMs that the difference in surprisal at the preferred and non-preferred condition is not significant ($p > 0.05$). The observation that surprisal estimates between the preferred and non-preferred conditions are similar provides evidence that higher surprisal estimates from LMs are not associated with processing less typical relations in non-preferred conditions, because if this were the case, there would have been significantly greater surprisal for the non-preferred condition compared to the preferred condition. This further supports that higher surprisal estimates from LMs could reflect costs specific to processing implicit meaning which is only present in the coerced condition out of the three conditions in Dataset 1.

\subsection{Experiment 2: Event/Neutral verb + Event/Entity NP}
\label{result: dataset2}
In this experiment, it is consistent across all five models that at the target region, Friedman's test of differences among repeated measures shows that there is significant difference (in all cases, $p < 0.001$) among surprisal estimates measured in the (a) Event verb + Event NP, (b) Neutral verb + Event NP, (c) Event verb + Entity NP, and (d) Neutral verb + Entity NP conditions.

\subsubsection{Greater surprisal for coerced condition or Entity NP?}
Consistent across all five models, post-hoc tests show that surprisal estimates at the target region are significantly higher for condition (c), which involves an implicit activity, compared to both conditions (a) and (b) ($p < 0.01$ for all pairwise comparisons). However, in comparing condition (c) to condition (d) which has no implicit activity, only DistilGPT-2 shows greater surprisal for the condition with implicit activity ($p < 0.05$), whereas no significant difference in surprisal estimates between these two conditions is observed for the four other models ($p > 0.05$). These results demonstrate that for the case of DistilGPT-2, surprisal estimates at the target region do seem to reflect effects associated with implicit meaning in the coerced condition, since higher surprisal estimates are specific to sentences with implicit activities compared to other sentences. However, for the other four models, the measurements suggest that the greater surprisal observed could potentially be attributed to a general effect due to processing entity-denoting NPs (conditions (c) and (d)). 

Further, post-hoc tests for the larger GPT-2 models (medium, large and xl) show a reliable effect of surprisal at the target region being greater for condition (d) compared to conditions (a) and (b) ($p < 0.05$) but conditions (c) and (d) are not significantly different from each other. This points towards greater surprisal reflecting processing cost associated with entity-denoting NP in general rather than the coerced condition. As all sentences in Dataset 1 had entity-denoting NPs, no comparison was made for different NPs and this effect was not observed from our experiments on Dataset 1. This new finding using Dataset 2 highlights that investigation of a phenomenon based on one setup or sentence template is not enough, and the need to have carefully varied controls.

\subsubsection{Greater surprisal $\neq$ less specific verbs}
If greater surprisal at the target region is only due to verbs like \textit{started} being (arguably) less specific than verbs like \textit{saw}, behavior within condition pairs (a) and (c), as well as (b) and (d) would be similar, and any difference would only be across the pairs where the surprisal for (a) and (c) would be greater than that of (b) and (d) respectively. We rule out this possibility because based on the post-hoc tests, surprisal for condition (c) is consistently greater than (a) across all five models, and surprisal for (d) is greater than (b) for the medium, large and xl models. There is also a reliable trend that surprisal estimates for conditions (a) and (b) are not significantly different, and surprisal for condition (c) is not significantly greater than (d) except for the case of DistilGPT-2. These analysis show that greater surprisal from LMs does not reflect processing cost due to verbs like \textit{started} being (arguably) less specific.

\subsubsection{Greater surprisal $\neq$ event interpretations of NPs}
If the difference in surprisal at the target region is generally associated with processing event interpretations of NPs rather than computing implicit meaning, LMs' surprisal for conditions (a), (b) and (c) where NPs are assigned event interpretations would each be significantly greater than condition (d) where the NP takes on an entity interpretation. This explanation is ruled out because surprisal estimates at conditions (a) and (b) are never higher than that at (d) in all cases, and surprisal for condition (c) is not significantly greater than (d) generally (except for the case of DistilGPT-2).

\subsection{Experiment 3: Coerced/Anomalous/Control}
Across all five models and the three different measure positions, Friedman's test show that there is a significant difference (in all cases, $p < 0.001$) among surprisal estimates measured in the coerced, anomalous and control conditions. At the differing verb and target region, it is consistent across all five models that post-hoc tests show that all pairwise differences are significant ($p < 0.05$). At the post-target region, it is consistent across the five models that surprisal for the anomalous condition is significantly higher ($p < 0.01$) than that of the coerced and control conditions.  

\label{result: dataset3}
\subsubsection{Greater surprisal for coerced condition compared to control condition}
At the differing verb and target region, there is a reliable trend of significantly higher surprisal estimates for the coerced condition compared to the control condition. This observation for the target region aligns with the results from measurements on Dataset 1, which also suggests that greater surprisal could reflect additional processing cost for dealing with implicit meaning that is present in the coerced condition.

\subsubsection{Even greater surprisal for processing anomalous condition}
However, we also found that across all five models and all three measure positions, there is consistently higher surprisal estimates for the anomalous condition compared to the coerced condition (and control condition). This raises the question of whether greater surprisal demonstrated by LMs is associated with reconstructing implicit meaning (in the coerced condition) or that it reflects anomaly detection. Given such observation, we cannot dismiss the possibility that LMs are processing coercion like a weaker form of anomaly.

\section{Related work}
\label{sec:related}
A range of prior work has contributed towards rigorously probing and improving LMs, including uncovering how LMs exploit common surface patterns \citep{TACL2887}, introducing probing method that considers utility of information by models \citep{TACL2423}, datasets that isolate specific linguistics phenomenon \citep{warstadt-etal-2020-blimp-benchmark}, strategies for improving performance on out-of-distribution examples \citep{TACL3499}, and for avoiding spurious correlations \citep{TACL2055}. To better evaluate LMs, \citet{TACL3539} use a smaller, carefully curated diagnostic dataset,  \citet {TACL2767} control for parameters highly correlated with the target phenomenon, \citet{Ettinger2019WhatBI} and \citet{TACL3197} draw insights from cognitive sciences. Our approach further contributes to such efforts. By making measurements at different critical regions and using targeted psycholinguistics diagnostic datasets to perform a series of connected experiments, we demonstrate the inadequacy of drawing conclusions based on each single probing experiment alone and the need to prioritize the depth of analysis.

Another line of relevant works study phenomena where meaning in language is not explicitly expressed, such as in idioms and similes \citep{TACL3505}, metaphor processing \citep{TACL1912}, relations between entities \citep{elazar-2021-tne}, as well as implicit reasoning strategies for question-answering \citep{TACL2641}, and sentence decontextualization \citep{TACL2667}. On the phenomenon of coercion, \citet{elazar-etal-2020-extraordinary} was the first to crowdsource data for it, but achieved low agreement scores between annotators (Fleiss' Kappa, $k = 0.24$) \citep{Fleiss1971MeasuringNS}, and the phenomenon has been otherwise left unexplored. Our work circumvents this issue by using diagnostics from psycholinguistic experiments, and fills a gap in this space.

\section{Discussion and conclusion}

This work presents a systematic study examining LMs' behavior when processing sentences with implicit meaning, using the complement coercion phenomenon, which has been largely left unexplored. Effects were observed at the target region rather than the minimally differing verb region, highlighting that differences in language can be more than what meets the eye and the importance of looking at multiple sentence regions critical for the phenomenon. Our series of experiments designed to ask targeted questions are effective in identifying potential confounds which would be otherwise mistaken for coercion effects. We arrive at a richer and more accurate analysis by making measurements at various critical regions in sentences (measure more) and having a series of targeted follow-up experiments (question more), methodologies that can be potentially applied to various future works dealing with language in general. We hope our work can inspire future efforts to measure more and question more.

\bibliography{tacl2021}
\bibliographystyle{acl_natbib}









  

\end{document}